\documentclass[10pt,twocolumn,letterpaper]{article}

\usepackage{cvpr}              
\usepackage[accsupp]{axessibility}  

\usepackage[dvipsnames]{xcolor}

\definecolor{cvprblue}{rgb}{0.21,0.49,0.74}
\usepackage[pagebackref,breaklinks,colorlinks,citecolor=cvprblue]{hyperref}
\usepackage{multirow}

\def\paperID{*****} 
\def\confName{CVPR}
\def\confYear{2024}

\title{MIPI 2024 Challenge on Few-shot RAW Image Denoising: Methods and Results}

\makeatletter
\def\thanks#1{\protected@xdef\@thanks{\@thanks
        \protect\footnotetext{#1}}}
\makeatother

\author{
\textbf{Challenge and Workshop Organizers}\\
Xin Jin$^1$\thanks{$^1$VCIP, CS, Nankai University}
\quad Chunle Guo$^1$ \quad 
Xiaoming Li$^2$\thanks{$^2$S-Lab, Nanyang Technological University} \quad Zongsheng Yue$^2$ \quad Chongyi Li$^1$ \quad Shangchen Zhou$^2$ \\
Ruicheng Feng$^2$ \quad Yuekun Dai$^2$ \quad Peiqing Yang$^2$  \quad  
Chen Change Loy$^2$ \\
\vspace{-5pt}
\\
\textbf{Challenge Participants}\\
Ruoqi Li$^3$\thanks{$^3$Video Algorithm Group, Camera Department, Xiaomi Inc., China} \quad Chang Liu$^3$ \quad Ziyi Wang$^3$ \quad Yao Du$^3$ \quad Jingjing Yang$^3$ \quad Long Bao$^3$ \\
Heng Sun$^3$ \quad 
Xiangyu Kong$^4$\thanks{$^4$Samsung Research China - Beijing (SRC-B)} \quad Xiaoxia Xing$^4$ \quad Jinlong Wu$^4$ \quad Yuanyang Xue$^4$ \quad Hyunhee Park$^5$\thanks{$^5$Department of Camera Innovation Group, Samsung Electronics} \\
Sejun Song$^5$ \quad Changho Kim$^5$ \quad Jingfan Tan$^6$\thanks{$^6$Sun Yat-sen University} \quad Wenhan Luo$^6$ \quad Zikun Liu$^4$ \quad
Mingde Qiao$^7$\thanks{$^7$Harbin Institute of Technology, China} \\
Junjun Jiang$^7$ \quad Kui Jiang$^7$ \quad Yao Xiao$^7$ \quad Chuyang Sun$^7$ \quad Jinhui Hu$^8$\thanks{$^8$Smart City Research Institute of China Electronics Technology Group Corporation} \quad Weijian Ruan$^8$ \\
Yubo Dong$^{9,10}$\thanks{$^9$Xidian University, Xi’an, China}\thanks{$^{10}$Nanyang Technological University, Singapore} \quad
Kai Chen$^{11}$\thanks{$^{11}$University of Electronic Science and Technology of China, China} \quad
Hyejeong Jo$^{12}$\thanks{$^{12}$National Hanbat University, Daejeon, South Korea} \quad
Jiahao Qin$^{13}$\thanks{$^{13}$North University of China, China} \quad Bingjie Han$^{13}$ \quad Pinle Qin$^{13}$ \\
Rui Chai$^{13}$ \quad Pengyuan Wang$^{13}$
\vspace{-15pt}
}

\begin{document}
\maketitle
\begin{abstract}
\vspace{-4mm}
The increasing demand for computational photography and imaging on mobile platforms has led to the widespread development and integration of advanced image sensors with novel algorithms in camera systems.
However, the scarcity of high-quality data for research and the rare opportunity for in-depth exchange of views from industry and academia constrain the development of mobile intelligent photography and imaging (MIPI).
Building on the achievements of the previous MIPI Workshops held at ECCV 2022 and CVPR 2023, we introduce our third MIPI challenge including three tracks focusing on novel image sensors and imaging algorithms.
In this paper, we summarize and review the Few-shot RAW Image Denoising track on MIPI 2024.
In total, 165 participants were successfully registered, and 7 teams submitted results in the final testing phase.
The developed solutions in this challenge achieved state-of-the-art performance on Few-shot RAW Image Denoising.
%
More details of this challenge and the link to the dataset can be found at \href{https://mipi-challenge.org/MIPI2024/}{https://mipi-challenge.org/MIPI2024}.
\end{abstract}    
\section{Introduction}
\label{sec:intro}
Noise, as an unavoidable part of the image capturing process, has recently been deeply explored by many researchers~\cite{buades2005non,zhang2017beyond,ulyanov2018deep,lehtinen2018noise2noise,abdelhamed2018high,chen2018sid,wei2021physics}. Mobile terminals, such as smartphones, smart cameras, and fixed and mobile cameras, are often plagued by noise. Noise not only affects people's visual perception but also usually impacts more high-level downstream tasks. Therefore, the development of denoising techniques is extremely important.

Since the pioneering work of SID~\cite{chen2018sid}, the use of deep neural networks for RAW image denoising has gradually come into view~\cite{zhang2017beyond,ulyanov2018deep,lehtinen2018noise2noise,abdelhamed2018high,jin2023dnf}. Existing methods can be divided into three categories: 
1) collecting a large-scale paired dataset for training denoising networks~\cite{ulyanov2018deep,abdelhamed2018high,chen2018sid}; 
2) first collecting datasets necessary for noise parameter calibration~\cite{wei2021physics,feng2022learnability}, then using the calibrated noise model to synthesize noisy-clean paired data and training the neural network; 
3) collecting a few-shot amount of paired data, adopting pre-training on synthetic data, followed by fine-tuning the network with the few-shot paired dataset~\cite{jiniccv23led}.
The collection of clean images is extremely difficult, typically involving multi-frame fusion~\cite{buades2005non}, long exposure with low ISO, or a combination of both. However, these collection methods are not capable of capturing outdoor or dynamic scenes, leading to a small domain or poor quality of training data, which naturally impacts the performance of neural networks.
Therefore, methods that train with paired data are gradually fading from view, while calibration-based approaches~\cite{wei2021physics} are becoming mainstream. This is because calibration-based methods can obtain noise parameters that are close to the real noise distribution, and using these noise parameters can generate infinite paired data without being limited by the scene.
These methods require perfect modeling of noise to address the domain gap between real and synthetic data, but this is usually challenging because many factors can affect the noise distribution: temperature, camera model, and even the lens can have an impact on noise.
Referring to LED~\cite{jiniccv23led}, the approach based on few-shot fine-tuning effectively bridges these two methods. It allows for pre-training with infinite synthetic data and fine-tuning with few-shot paired data, where the collection cost of few-shot data is extremely low.
We encourage researchers to use pre-training and few-shot fine-tuning methods to train neural networks for RAW image denoising. This can not only reduce the demand for paired data but also solve the domain gap between the noise model and real noise to a certain extent.
In response to the growing demand among smartphone and camera manufacturers, this competition focuses on developing raw image denoising methods with insufficient data.
Participants can use publicly available datasets~\cite{chen2018learning,wei2021physics} for pre-training and fine-tuning with data from two different camera manufacturers that we will provide.

We hold this challenge in conjunction with the third MIPI Challenge which will be held on CVPR 2024. Similar to the previous MIPI challenge~\cite{sun2023mipi,sun2023mipi2,dai2023mipi,zhu2023mipi}, we are seeking an efficient and high-performance image restoration algorithm to be used for raw image denoising with insufficient data. MIPI 2024 consists of three competition tracks:

\begin{itemize}
    \item \textbf{Few-shot RAW Image Denoising} is geared towards training neural networks for raw image denoising in scenarios where paired data is limited.
    \item \textbf{Demosaic for HybridEVS Camera} is to reconstruct HybridEVS's raw data which contains event pixels and defect pixels into RGB images.
    \item \textbf{Nighttime Flare Removal} is to improve nighttime image
    quality by removing lens flare effects.
\end{itemize}

\section{MIPI 2024 Few-shot RAW Image Denoising}
\label{sec:track}

To promote the development of RAW image denoising and gradually shift from training methods based on paired data or noise model calibration to a training strategy utilizing few-shot fine-tuning.
This challenge aims to advance research on few-shot RAW image denoising.

\subsection{Datasets}
In this competition, we collected paired data from 30 different scenes using cameras from two different manufacturers. Each scene includes two different ISO settings, five different brightness settings, and for each setting, 5 shots were taken. 
To obtain clean RAW images as ground truth, we employed multi-frame fusion denoising during the process of creating the ground truth. One camera model was used to train on data from two scenes, while the other model was used to train on data from five scenes. Both camera models used data from five scenes for validation and were tested on the remaining scenes. Participants can use these data for fine-tuning or directly train on these data. Additionally, we provide a simple image signal processing pipeline for participants to visualize the denoised RAW data. The final RGB data is used for evaluation.
Since we have not disclosed the camera models, participants cannot use the data for training through calibration methods. However, participants can use publicly available RAW datasets for pre-training. 
Our dataset provides 1920$\times$1920 RAW patches for fine-tuning and validation, and all RAW data are in the Bayer pattern.

\subsection{Evaluation}
We employ the standard Peak Signal To Noise Ratio (PSNR) and the Structural Similarity Index (SSIM) for evaluation. SSIM is calculated in grayscale, as is commonly used in the literature. 
To integrate two evaluation metrics on the same scale, we adopted the following approach:
\begin{equation}
    Score=\log_k(SSIM\times k^{PSNR}),
\end{equation}
which equals to
\begin{equation}
    Score=PSNR+\log_k(SSIM).
\end{equation}
In our implementation, $k=1.2$.
Participants can view these metrics of their submission to optimize the model's performance.

\subsection{Challenge Phase}
The challenge consisted of the following phases:
\begin{enumerate}
    \item Development: The registered participants get access to the data and baseline code, and are able to train the models and evaluate their methods locally (only with the train split).
    \item Validation: The participants can upload their models to the remote server to check the fidelity scores on the validation dataset, and to compare their results on the validation leaderboard.
    \item Testing: The participants submit their final results, code, models, and factsheets.
\end{enumerate}
\section{Challenge Results}

Among $165$ registered participants, $7$ teams successfully submitted their results, code, and factsheets in the final test phase.
Table \ref{tab:result} reports the final test results and rankings of the teams. 
The methods evaluated in Table \ref{tab:result} are briefly
described in Section \ref{sec:methods} and the team members are listed in Appendix.
Finally, the MiAlgo\_AI team is the first place winner of this challenge, while  BigGuy team win the second place and SFNet-FR team is the third place, respectively.

\begin{table}[]\small
\centering
\caption{Results of MIPI 2024 challenge on few-shot RAW image denoising. Best, second and third are denoted in {\color{red}{red}}, {\color{blue}{blue}} and \textbf{bold}, respectively.}
\label{tab:result}
\begin{tabular}{l|cc|c}\toprule
{\textbf{Team name}}                                                      & PSNR   & SSIM   & Score   \\\midrule
MiVideoNR                                                                 & \color{red}{$31.225_{(\textbf{1})}$} & \color{blue}{$0.9542_{(\textbf{2})}$} & \color{red}{$30.9645_{(\textbf{1})}$} \\
Samsung$\dagger$                                                          & \color{blue}{$30.965_{(\textbf{2})}$} & \color{red}{$0.9545_{(\textbf{1})}$} & \color{blue}{$30.7064_{(\textbf{2})}$} \\
AIIA                                                                      & $\textbf{29.981}_{(\textbf{3})}$ & $\textbf{0.9487}_{(\textbf{3})}$ & $\textbf{29.6886}_{(\textbf{3})}$ \\
MS-Denoimer                                                               & $29.926_{(\textbf{5})}$ & $0.9481_{(\textbf{4})}$ & $29.6300_{(\textbf{4})}$ \\
MyTurn                                                                    & $29.926_{(\textbf{4})}$ & $0.9447_{(\textbf{6})}$ & $29.6096_{(\textbf{5})}$ \\
HBNU                                                                      & $29.734_{(\textbf{6})}$ & $0.9462_{(\textbf{5})}$ & $29.4263_{(\textbf{6})}$ \\
Erlong Mountain Team                                                      & $23.433_{(\textbf{7})}$ & $0.9121_{(\textbf{7})}$ & $22.9211_{(\textbf{7})}$ \\\bottomrule
\multicolumn{4}{l}{$\dagger$\shortstack{\fontsize{5.3pt}{8pt}\selectfont{Full name: Samsung MX(Mobile eXperience) Business \& Samsung Research China - Beijing (SRC-B)}}}
\end{tabular}
\end{table}

\section{Methods}
\label{sec:methods}

\paragraph{\bf MiVideoNR}

\begin{figure}[t]
    \centering
     \includegraphics[width=0.48\textwidth]{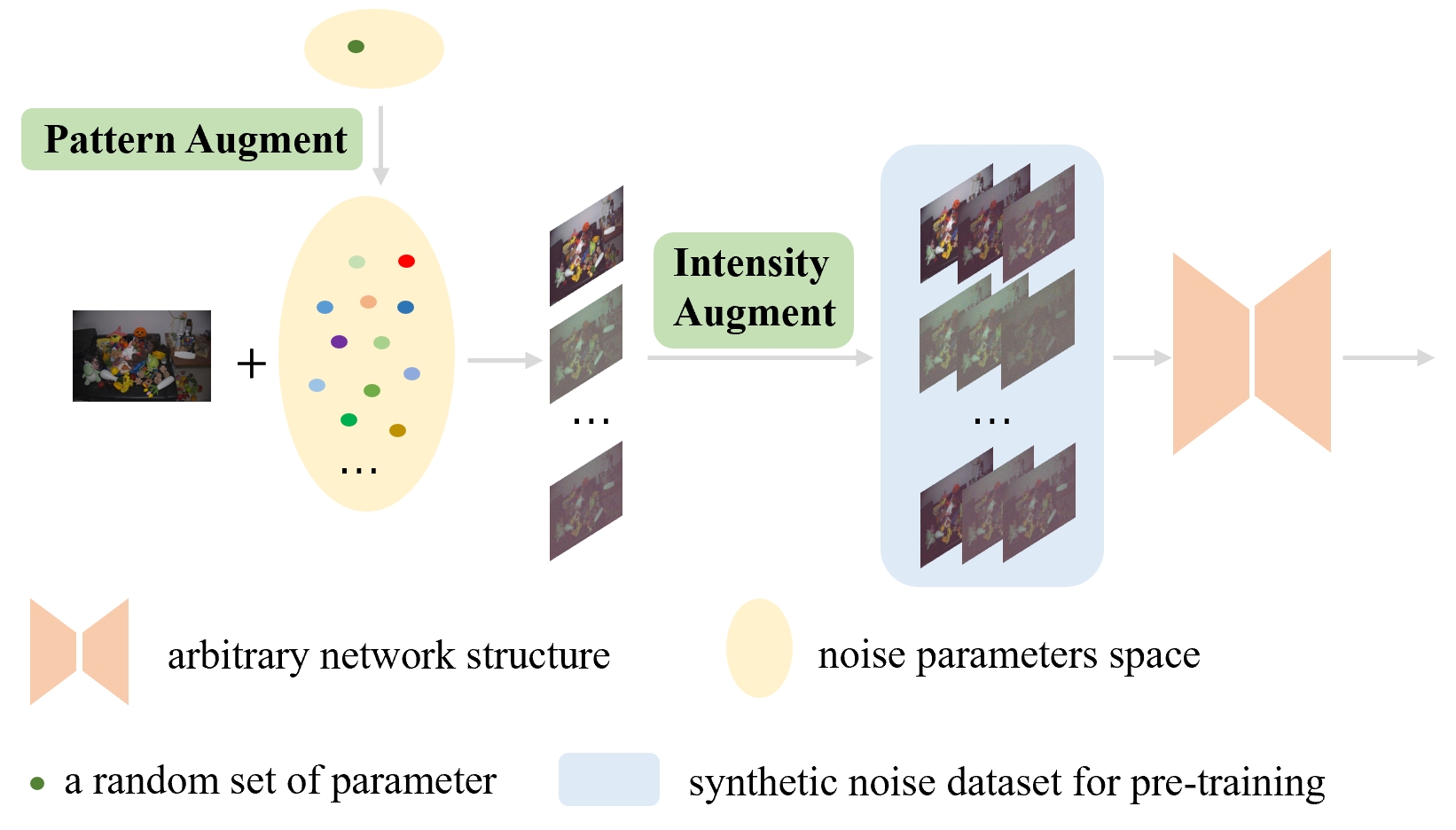}
    \caption{Illustration of the pre-train stage of team MiVideoNR.}
    \label{fig:mi}
    \vspace{-0.5cm}
\end{figure}

This team proposes a calibration-free two-stage pipeline for learning from synthetic noise to real noise. 
In the first stage, the authors pre-train NAFNet~\cite{chu2022nafssr} from scratch using an enormous synthetic noise set. They use high-quality clean images from two DSLR datasets~\cite{fivek, realsr} and add noise modeling from Possion-Gaussian distribution to form training samples. In order to enlarge the randomness and diversity of the noise parameters, the authors carefully design a pattern-augment module to disturb the initial noise parameter and an intensity-augment module to amplify the noise by different ratios (see Figure~\ref{fig:mi}), which enhance the network with strong generalization ability for efficiently adapting to specific camera sensor. This stage is train with the batch size 8 and patch size 224 for around 300k iterations. The model is optimized by AdamW optimizer using L1 loss, with the initial learning rate of $3e-4$, which decreases by 0.6 in 100k and 200k iterations.
In the second stage, the authors use the provided paired real data to fine-tune the network without any modifications of the structures, with the batch size 4 and patch size 640 for around 5k iterations. The model is optimized by AdamW optimizer using L1 loss, with the initial learning rate of $1e-4$ and decreases by 0.6 in 1k and 3k iterations.
To further strengthen the generalization ability and boost the performance in real-world scenarios, the authors combine the predictions of four models with different image entry characteristics (\emph{i.e.} different input resolutions and whether to keep the negative value of the noise data) to obtain more robust denoising results. 

\paragraph{\bf Samsung\protect\footnote{{\fontsize{5.3pt}{8pt}\selectfont{Full name: Samsung MX(Mobile eXperience) Business \& Samsung Research China - Beijing (SRC-B)}}}}

\begin{figure*}[t]
  \centering
  \includegraphics[width=0.95\linewidth]{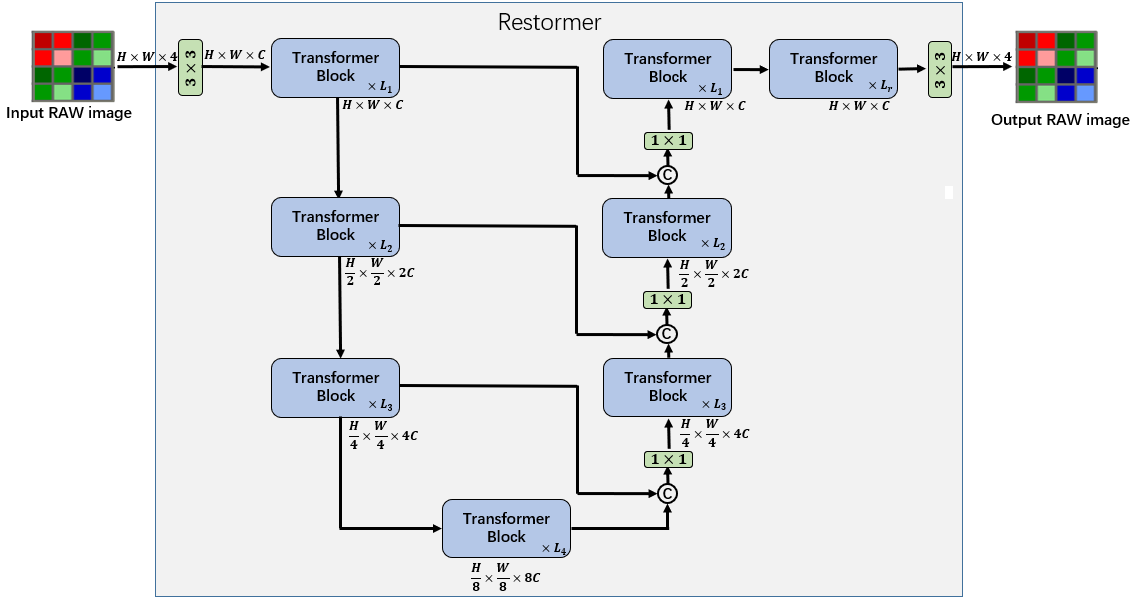}
  \caption{Model Framework of Samsung}
  \label{fig:Samsungframwork}
\end{figure*}

We propose to train a powerful pre-trained model using real data from different cameras to solve the weak generalization ability of the model on unseen data. Our model architecture is basically the same as the Restormer~\cite{zamir2022restormer} model as shown in Fig.~\ref{fig:Samsungframwork}, but with 4 channels as input and output as our task is in raw space instead of sRGB space. The training phase is comprised of two main processes: pre-training and fine-tuning. For pre-training stage, to train a model with strong generalization, we collected a dataset with more than 1500 pairs using our DSLR cameras as described in the paper~\cite{wei2021physics} to pre-train model. However, we observed that there is a color difference between the prediction and GT during the pre-training. To solve the color bias issue, we design a color loss as shown below:
\[
L_c = \lvert\frac{\hat{y}}{\hat{M}}-\frac{y}{M}\rvert + \lvert \hat{M} - M \rvert,
\]
in which the $L_c$ is color loss, $\hat{y}$ is the predicted pixel, $y$ is the GT, $\hat{M}$ and $M$ are the two mean values of all the 4 channels of the predicted and the ground truth respectively. In the final, our loss is like below:
\[
L=L_1 + \alpha L_c,
\]
in which we set $\alpha =0.5$, but will adjust it during the training process to balance the color loss and the $L_1$ loss. For fine-tuning, the model only uses $L_1$ loss to train on the training dataset provided by the organizer. Concretely, we use AdamW~\cite{decoupled} optimizer ($\beta_1=0.9$, $\beta_2=0.999$, weight decay $0.0001$) with the cosine annealing strategy, where the learning rate gradually decreases from the initial learning rate $5\times10^{-5}$ to $1\times10^{-7}$ for $5\times10^5$ iterations in pre-training stage while the learning rate gradually decreases from the initial learning rate $1\times10^{-6}$ to $1\times10^{-7}$ for $2\times10^5$ iterations in fine-tuning stage. The training batch size is set to $20$ and patch size is $180$. Horizontal/vertical flipping and rotation are used for data augmentation. All experiments are conducted on A100 GPU. To further improve performance of result during inference, self-ensemble is employed to produce the final results. Specifically, rotation and horizontal flipping are applied to the input images to obtain a set of images. These images are then input into our model. The final prediction is made by taking the mean of these outputs.

\begin{figure*}[t]
    \centering
     \includegraphics[width=0.88\textwidth]{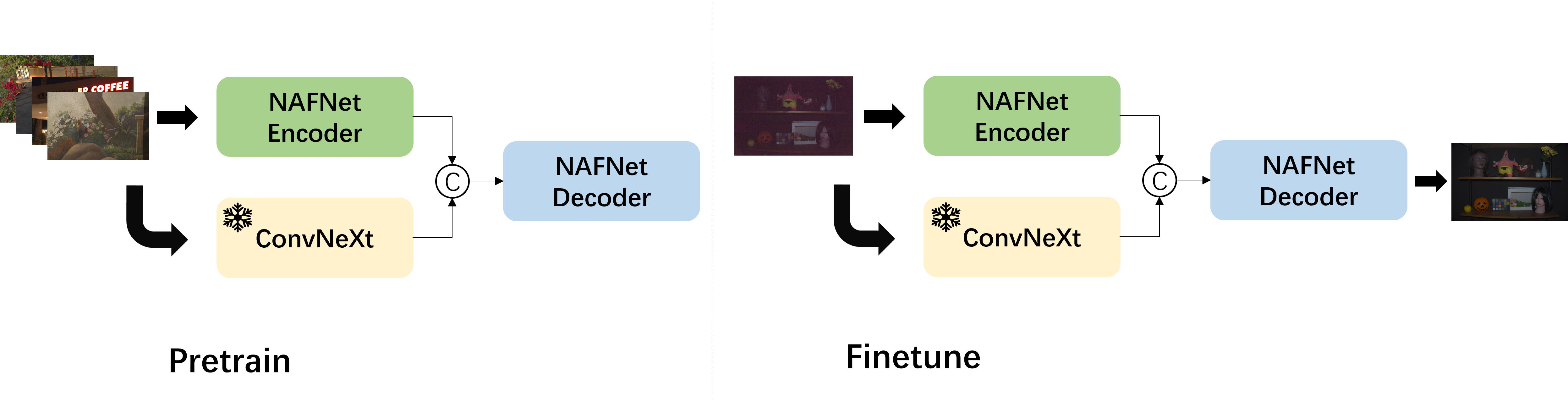}
    \caption{Illustration of the two stages of team AIIA}
    \label{fig:AIIA}
    \vspace{-0.5cm}
\end{figure*}

\paragraph{\bf AIIA}
Due to the characteristics of few-shot tasks, insufficient data often has a certain impact on model effectiveness. For higher-level tasks, pretrained models are commonly integrated into the network to enhance performance in downstream tasks. Leveraging this approach, we integrate the ConvNeXt~\cite{liu2022convnet} network, which has been pretrained on the ImageNet~\cite{5206848} dataset, into our network architecture. The primary aim is to leverage the powerful modeling capabilities demonstrated by this structure on large-scale datasets to strengthen our network's feature extraction and denoising abilities, thereby improving the model's generalization capability. This method not only utilizes pretrained ConvNeXt's efficient feature extraction capabilities but also enhances the model's denoising performance.The specific work is as follows(see Figure~\ref{fig:AIIA}): In the first phase, we add a prior branch to the NAFNet~\cite{chu2022nafssr} network and pretrain it on the SID~\cite{chen2018sid} dataset using five virtual cameras\cite{Wei_Fu_Zheng_Yang_2021}. In the second phase, we fine-tune the network using paired real data, following a process similar to that used in LED\cite{jiniccv23led}. The model employs an L1 loss and is optimized with Adam, trained during the pretraining phase with a learning rate that decreases from $2e-4$ to $0$ over 289,800 iterations. In the fine-tuning phase, we significantly increase the number of iterations to 45,000, with the learning rate decreasing from $2e-4$ to $0$. Finally, in order to further improve the denoising results, we first flip and rotate the images as the input to the model, then restore and average the outputs.

\begin{figure*}[t]
    \centering
     \includegraphics[width=0.85\textwidth]{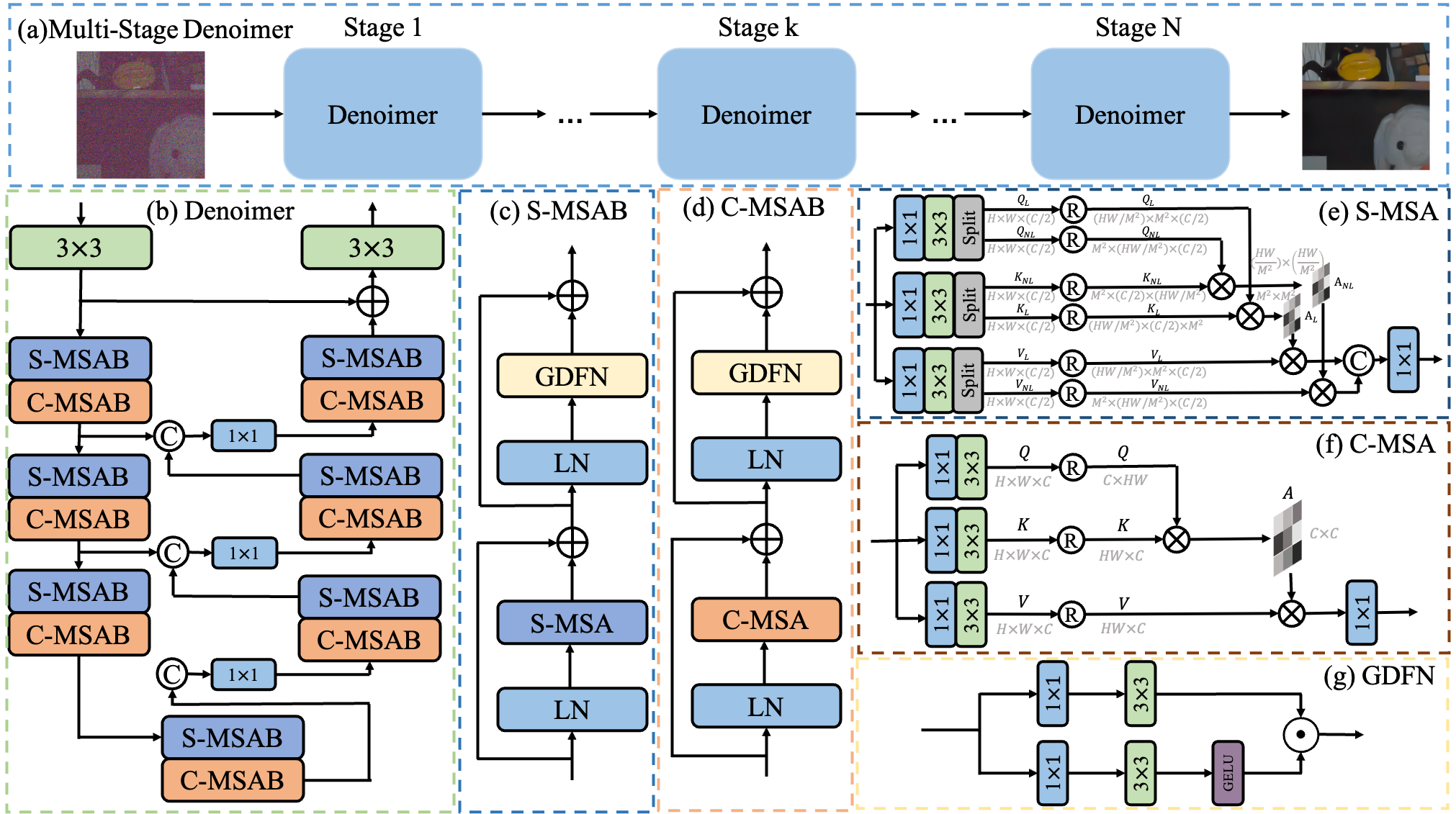}
     \caption{Diagram of MS-Denoimer. (a) The diagram of the Multi-Stage Denoimer. (b) The diagram of the single-stage Denoimer. (c) The diagram of the Spatial Multi-head Self-attention Block (S-MSAB). (d) The diagram of the Channel-wise Multi-head Self-attention Block (C-MSAB). (e) The illustration of the Spatial Multi-head Self-Attention (S-MSA). (f) The illustration of the Channel-wise Multi-head Self-Attention (C-MSA). (g) The illustration of the Gated-DConv Feedforward Network (GDFN).}
    \label{fig:MS-Denoimer}
    \vspace{-0.cm}
\end{figure*}

\paragraph{\bf MS-Denoimer}
{
This team proposes a Multi-Stage Denoiser with Spatial and Channel-wise Attention (MS-Denoimer, see Figure \ref{fig:MS-Denoimer}) for raw image denoising. Specifically, it utilizes a local and non-local multi-head self-attention mechanism \cite{dauhst, dernn_lnlt} to capture spatial correlations and a channel-wise multi-head self-attention mechanism \cite{mst, zamir2022restormer, rdluf_mixs2} to address channel-specific dependencies. These elements compose basic units: Spatial Multi-head Self-Attention Blocks (S-MSABs) and Channel-wise Multi-head Self-Attention Blocks (C-MSABs), which are alternately built up the single-stage Denoising Transformer (Denoimer). The Denoimer exploits a U-shaped structure to extract multi-resolution contextual information. Finally, the MS-Denoimer, cascaded by several Denoimers, progressively improves the reconstruction quality from coarse to fine. For training,  the authors pre-train MS-Denoimer on the SID \cite{chen2018sid} dataset and finetune it with Camera1 and Camera2 train sets. The objective of the model was to minimize the Charbonnier loss.  The authors employ the Adam optimizer with hyperparameters $\beta_1 = 0.9$  and $\beta_2 = 0.999$.  The pre-training learning rate is 4e-4 and the fine-tuning learning rate is 1e-4. The pre-training and fine-tuning process spanned 300 epochs and 200 epochs, respectively. The cosine annealing scheduler with linear warm-up was utilized. In the final inference stage, the authors integrate test time augmentation and model ensemble, including horizontal flip, vertical flip, and 90-degree rotation. 
}

\begin{figure}[t]
    \centering
     \includegraphics[width=0.45\textwidth]{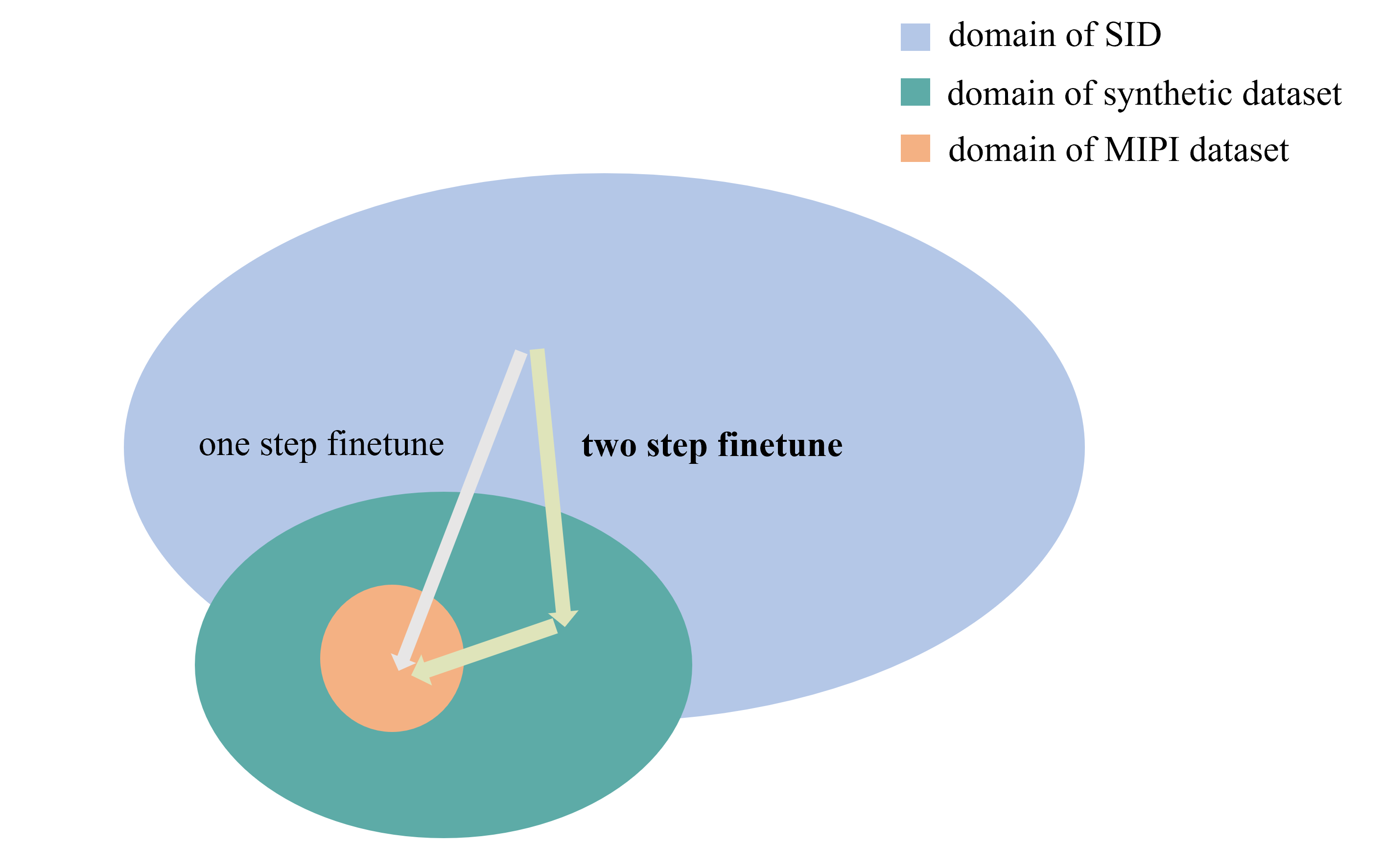}
    \caption{Illustration of the two step finetune strategy of team MyTurn}
    \label{fig:MyTurn}
    \vspace{-0.5cm}
\end{figure}

\paragraph{\bf MyTurn}
{
In the low-level domain, the U-Net architecture has been widely used. This team modifies the NAFNet \cite{chu2022nafssr}, which is based on the U-Net architecture, by increasing the number of blocks in the encoder as well as the network's channel count. The authors divide the training into three stages. In the first stage, the authors pretrain the model on the SID-Sony A7S2 dataset \cite{chen2018sid}. In the second stage, the authors model the noise as a heteroscedastic Gaussian distribution. For each image under a camera id, the authors perform linear regression between the clean image and the noise to obtain the noise parameters . Then the authors finetune the model on a noise dataset synthesized based on MIPI dataset to narrow down the domain gap(see Figure~\ref{fig:MyTurn}). Finally, in the third stage, the author performs the last fine-tuning on the MIPI dataset. Through two fine-tuning operations, neural networks can more easily learn the distribution of noise information. For training, the authors randomly cropped 384 × 384 patches from the training images as inputs. The model is trained with L1 Charbonnier loss \cite{wu2017srpgan}. The learning rate is initialized as $1 \times 10^{-4}$, and gradually reduces to $1 \times 10^{-7}$ with the cosine annealing. In the final inference stage, the authors use self-ensemble strategy to get eight images. These images output will be averaged to get the final result. 
}
\paragraph{\bf HBNU}
{
Fine-tuning is widely utilized across various fields due to its ability to achieve remarkable performance even with a small amount of data. This team introduced a denoising architecture iteratively fine-tuning, leveraging the LED \cite{jiniccv23led}. During the training phase, we initially performed fine-tuning of the dataset provided by the MIPI competition with a pre-trained model using the training strategy of PMN \cite{Feng_Wang_Wang_Huang_2022} and the ELD \cite{Wei_Fu_Zheng_Yang_2021} dataset. The fine-tuned model was then used to iteratively fine-tune the dataset provided by the MIPI competition. The proposed method involves initially utilizing a pre-trained model from a different domain to obtain features, but through iterative fine-tuning, the model can better understand and incorporate the characteristics of specific tasks or domains. Additionally, the model becomes more suitable for specific tasks, alleviating issues of overfitting.
\begin{figure}[!ht]
    \centering
     \includegraphics[width=0.45\textwidth]{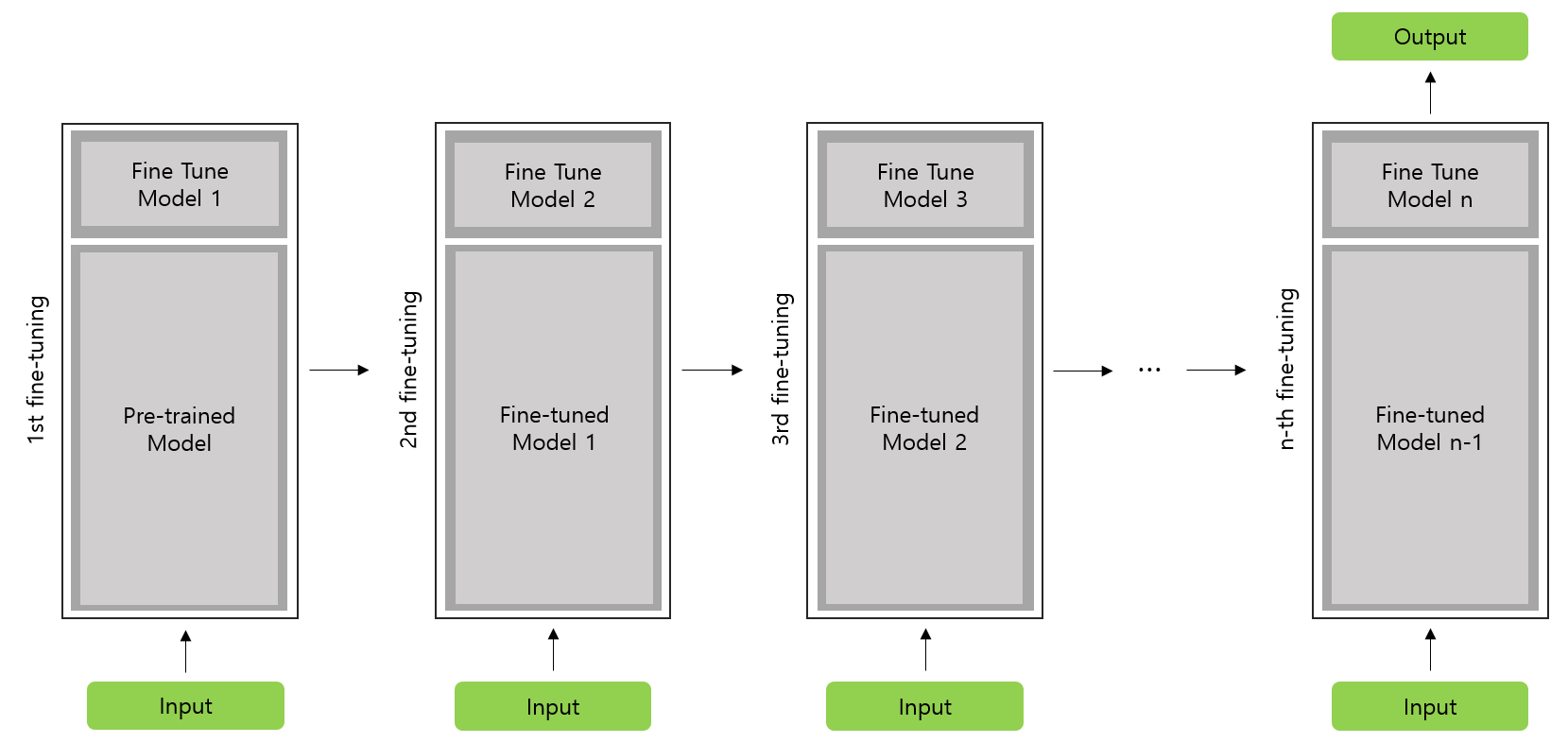}
    \caption{Illustration of the fine-tuning iteration proposed by the HBNU team.}
    \label{fig:HBNU}
    \vspace{-0.5cm}
\end{figure}
During the testing phase, Test-Time Data Augmentation \cite{timofte2016seven} was applied to enhance the predictions of the image. As described in \cite{timofte2016seven}, The original image, along with images rotated by 90°, 180°, and 270°, as well as the vertically flipped image and its rotations by 90°, 180°, and 270°, totaling 8 images, are denoising and then averaging. This team training dataset by randomly cropping a 512 x 512 region. The model utilized the Adam optimizer and underwent 1000 iterations. The learning rate was fixed at \(10^{-4}\) during the training. Finally, the OMNR branch was trained for 500 iterations with a learning rate of \(10^{-5}\).
}
\paragraph{\bf Erlong Mountain Team}
{
U-Net has been widely used in many application fields due to its excellent image segmentation ability and efficient network structure.The team proposed a denoising network based on UNet++\cite{zhou2019unet++} in a low-light environment. By integrating the UNet++ network with the CBAM \cite{101007} module, we ensure that the deep semantic information in the image is learned, which preserves the original information and effectively removes the noise present in the image. The authors divide the training into two phases. The first phase blends the datasets provided by the MIPI competition with the ELD \cite{Wei_Fu_Zheng_Yang_2021} and SID\cite{chen2018sid} datasets and augments their data, increasing the number of datasets ensures the robustness of the network and prevents overfitting during training (see Figure~\ref{fig:erlong}). In the second stage, the authors will train a hybrid dataset on the network, and the input images will be processed by the Unet++ network model, and finally the denoising will be obtained to obtain a clear image. Specifically, the network model contains four CBAM modules that process feature maps through channel attention and spatial attention after each downsampling step of UNet to enhance their discrimination while preserving important spatial information. This ensemble approach combines the CBAM module with the structure of the UNet model, allowing the UNet model to benefit from the CBAM module and improve its feature extraction and representation capabilities.

\begin{figure}[!ht]
    \centering
     \includegraphics[width=0.45\textwidth]{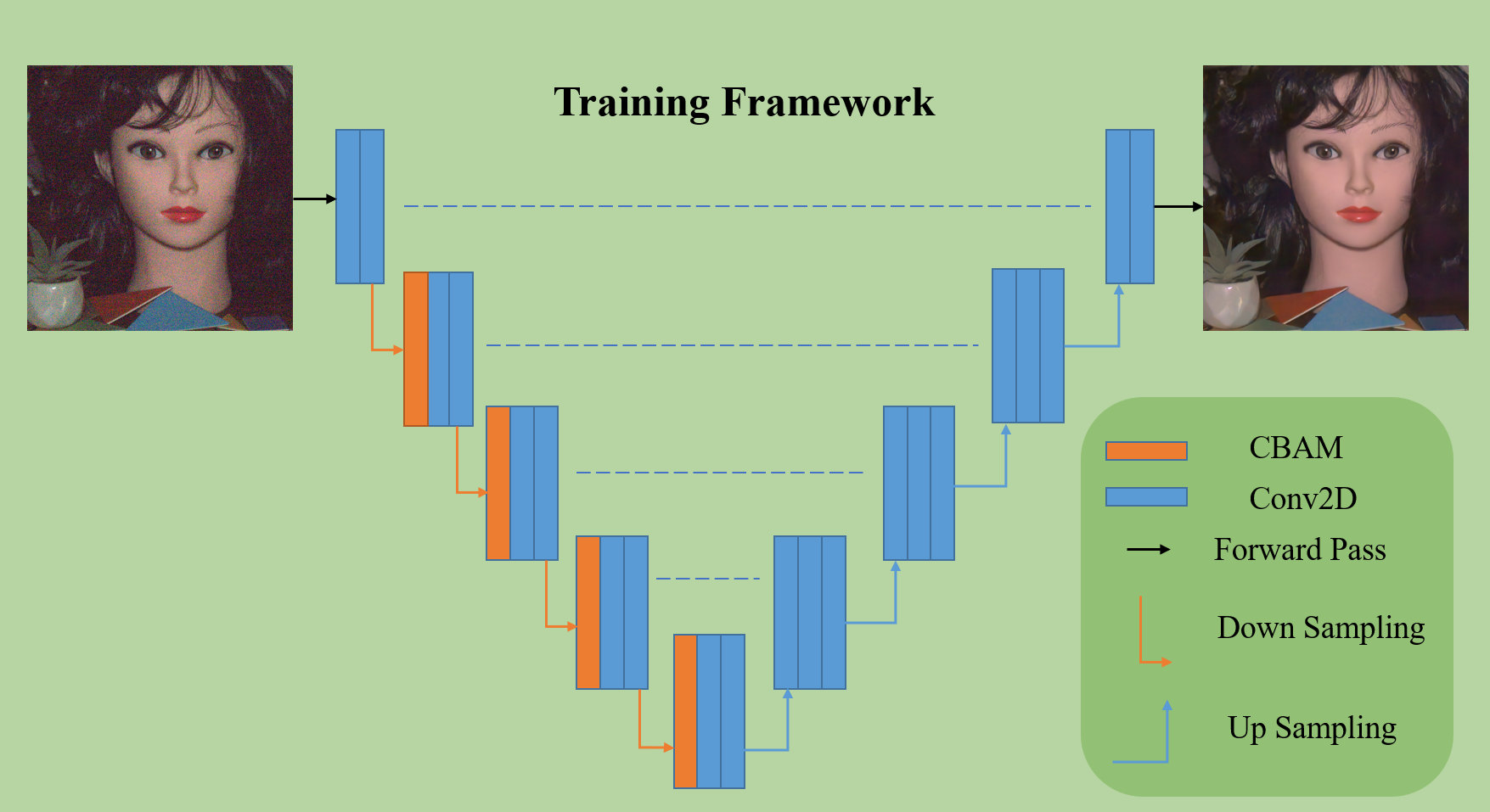}
    \caption{Illustration of the training framework of team Erlong Mountain Team.}
    \label{fig:erlong}
    \vspace{-0.5cm}
\end{figure}

Before conducting network training, we drew on the noise modeling methods used in ELD \cite{Wei_Fu_Zheng_Yang_2021} and the data augmentation methods in PRDNM\cite{Feng_Wang_Wang_Huang_2022}, with the aim of establishing more accurate noise models in low light environments and further removing noise.
For training, the authors randomly cropped a 512 x 512 region from the training set and performed random inversion/rotation enhancement data.
The model is trained for 1800 epochs using the Adam optimizer and a combination of SmoothLoss and MS-SSIM as the loss function. The base learning rate is set to \(2 \times 10^{-4}\) and the minimum learning rate is set to \(10^{-5}\). The optimizer restarts every 600 epochs and the learning rate is halved on restarts.
}
\section{Conclusions}
In this report, we review and summarize the methods and results of MIPI 2024
challenge on Few-shot RAW Image Denoising.

The MIPI 2024 Few-shot RAW Image Denoising Challenge aimed to advance the state-of-the-art in image denoising by focusing on few-shot learning techniques, accommodating the scarcity of paired data. The competition utilized datasets from diverse scenes captured with different camera manufacturers under varying ISO and brightness settings, specifically designed to challenge participants in training robust denoising models. The evaluation was based on PSNR and SSIM metrics, integrated through a unique scoring formula to balance the two. The challenge proceeded through development, validation, and testing phases, allowing participants to refine their models with a structured approach to tackling few-shot denoising tasks. Out of 165 registered teams, 7 made it to the final submission, showcasing a range of innovative solutions. The winning team, MiVideoNR, demonstrated superior performance in both PSNR and the combined score, closely followed by Samsung and AIIA, indicating a competitive and high-caliber field of entries. The diversity of approaches and the close competition underscores the challenge's success in pushing forward the boundaries of few-shot RAW image denoising research.

As for the methods, various teams introduced advanced and diverse methodologies for raw image denoising, demonstrating the field's innovative approaches to overcoming limitations posed by insufficient data. Despite their differences, these methods share common themes and strategies:

\begin{itemize}
    \item {\textbf{Pre-training and Fine-tuning Strategy}}: All teams have widely adopted strategies to overcome limited real-world data, utilizing extensive pre-training on synthetic or large datasets followed by fine-tuning on task-specific data.
    \item {\textbf{Synthetic to Real Noise Transition}}: Besides the naive pre-training and fine-tuning strategy, several teams developed methods to bridge the gap between synthetic noise and real-world noise conditions, enhancing the realism and diversity of training samples.
    \item {\textbf{Domain-specific Challenges}}: Specialized techniques to address challenges like color fidelity and low-light conditions, demonstrating a tailored approach to noise characteristics.
    \item {\textbf{Ensemble and Augmentation Techniques}}: Use of ensemble methods and test-time augmentation to improve model robustness and performance across varied noise conditions.
    \item {\textbf{DNN and Attention Mechanisms}}: All teams utilized state-of-the-art neural architectures, including NAFNet~\cite{chu2022nafssr}, Restormer~\cite{zamir2022restormer}, and U-Net variants~\cite{Uformer}, integrating spatial and channel-wise attention mechanisms for enhanced feature extraction and denoising performance.
\end{itemize}

\section{Teams and Affiliations}
\label{append:teams}

\subsection*{\bf MiVideoNR}
\noindent
{\bf Title:} From Synthetic to Real: A Calibration-free Pipeline for Few-shot Raw Image Denoising\\
{\bf Members:}\\
Ruoqi Li$^1$ (\href{liruoqi@xiaomi.com}{liruoqi@xiaomi.com})\\
Chang Liu$^1$\quad Ziyi Wang$^1$\quad Yao Du$^1$\quad Jingjing Yang$^1$\quad Long Bao$^1$\quad Heng Sun$^1$\\
{\bf Affiliations:}\\
$^1$ Video Algorithm Group, Camera Department, Xiaomi Inc., China

\subsection*{\bf Samsung MX(Mobile eXperience) Business \& Samsung Research China - Beijing (SRC-B)}
\noindent
{\bf Title:} Restormer based Few Shot RAW Denoising \\
{\bf Members:}\\
Xiangyu Kong$^1$(\href{xiangyu.kong@samsung.com}{xiangyu.kong@samsung.com})\\
Xiaoxia Xing$^1$\quad Jinlong Wu$^1$\quad Yuanyang Xue$^1$\quad Jingfan Tan$^3$\quad 
Zikun Liu$^1$\quad
Hyunhee Park$^2$\quad Sejun Song$^2$\quad Changho Kim$^2$\quad
Wenhan Luo$^3$\\
{\bf Affiliations:}\\
$^1$ Samsung Research China - Beijing (SRC-B)\\
$^2$ Department of Camera Innovation Group, Samsung Electronics\\
$^3$ Sun Yat-sen University

\subsection*{\bf AIIA}
\noindent
{\bf Title:} Transfering Pretrained Model for Few-shot Raw Image Denoising\\
{\bf Team Leader:}\\
MingDe Qiao$^1$ (\href{mailto:mingdeqiao@gmail.com}{mingdeqiao@gmail.com})\\
{\bf Members:}\\
Junjun Jiang$^1$\quad Kui Jiang$^1$\quad Yao Xiao$^1$\quad Chuyang Sun$^1$\quad Jinhui Hu$^2$\quad Weijian Ruan$^2$\\
{\bf Affiliations:}\\
$^1$ Harbin Institute of Technology, China\\
$^2$ Smart City Research Institute of China Electronics Technology Group Corporation

\subsection*{\bf MS-Denoimer}
\noindent
{\bf Title:} MS-Denoimer: Multi-Stage Denoiser with Spatial and Channel-wise Attention for Raw Image Denoising\\
{\bf Members:}\\
Yubo Dong (\href{ybdong@stu.xidian.edu.cn}{ybdong@stu.xidian.edu.cn})\\
{\bf Affiliations:}\\
$^1$ Xidian University, Xi'an, China \\
$^2$ Nanyang Technological University, Singapore 

\subsection*{\bf MyTurn}
\noindent
{\bf Title:} Domain Adaption Network\\
{\bf Members:}\\
Kai Chen (\href{chen\_kai@stu.xidian.edu.cn}{chen\_kai@stu.xidian.edu.cn})\\
{\bf Affiliations:}\\
University of Electronic Science and Technology of China, China
\subsection*{\bf HBNU}
\noindent
{\bf Title:} Enhancing Few-shot RAW Image Denoising through Iterative Fine-tuning and Test Time Data Augmentation \\
{\bf Members:}\\
Hyejeong Jo$^1$ (\href{jhjmain27@gmail.com}{jhjmain27@gmail.com})\\
Huiwon Gwon$^1$\quad Sunhee Jo$^1$\\
{\bf Affiliations:}\\
$^1$ National Hanbat University, Daejeon, South Korea
\subsection*{\bf Erlong Mountain Team}
\noindent
{\bf Title:} DeUnet: An Unet++ Network for Low-light Denoising\\
{\bf Members:}\\
Jiahao Qin$^1$ (\href{q656673477@gmail.com}{q656673477@gmail.com})\\
Bingjie Han$^1$\quad Pinle Qin$^1$\quad Rui Chai$^1$\quad Pengyuan Wang$^1$\\
{\bf Affiliations:}\\
$^1$ North University of China, China\\

{
    \small
    \bibliographystyle{ieeenat_fullname}
    \bibliography{main}

\begin{thebibliography}{33}
\providecommand{\natexlab}[1]{#1}
\providecommand{\url}[1]{\texttt{#1}}
\expandafter\ifx\csname urlstyle\endcsname\relax
  \providecommand{\doi}[1]{doi: #1}\else
  \providecommand{\doi}{doi: \begingroup \urlstyle{rm}\Url}\fi

\bibitem[Abdelhamed et~al.(2018)Abdelhamed, Lin, and Brown]{abdelhamed2018high}
Abdelrahman Abdelhamed, Stephen Lin, and Michael~S Brown.
\newblock A high-quality denoising dataset for smartphone cameras.
\newblock In \emph{CVPR}, 2018.

\bibitem[Buades et~al.(2005)Buades, Coll, and Morel]{buades2005non}
Antoni Buades, Bartomeu Coll, and J-M Morel.
\newblock A non-local algorithm for image denoising.
\newblock In \emph{CVPR}, 2005.

\bibitem[Bychkovsky et~al.(2011)Bychkovsky, Paris, Chan, and Durand]{fivek}
Vladimir Bychkovsky, Sylvain Paris, Eric Chan, and Fr{\'e}do Durand.
\newblock Learning photographic global tonal adjustment with a database of
  input / output image pairs.
\newblock In \emph{The Twenty-Fourth IEEE Conference on Computer Vision and
  Pattern Recognition}, 2011.

\bibitem[Cai et~al.(2019)Cai, Zeng, Yong, Cao, and Zhang]{realsr}
Jianrui Cai, Hui Zeng, Hongwei Yong, Zisheng Cao, and Lei Zhang.
\newblock Toward real-world single image super-resolution: A new benchmark and
  a new model.
\newblock In \emph{Proceedings of the IEEE International Conference on Computer
  Vision}, 2019.

\bibitem[Cai et~al.(2022{\natexlab{a}})Cai, Lin, Hu, Wang, Yuan, Zhang,
  Timofte, and Van~Gool]{mst}
Yuanhao Cai, Jing Lin, Xiaowan Hu, Haoqian Wang, Xin Yuan, Yulun Zhang, Radu
  Timofte, and Luc Van~Gool.
\newblock Mask-guided spectral-wise transformer for efficient hyperspectral
  image reconstruction.
\newblock In \emph{Proceedings of the IEEE/CVF Conference on Computer Vision
  and Pattern Recognition}, pages 17502--17511, 2022{\natexlab{a}}.

\bibitem[Cai et~al.(2022{\natexlab{b}})Cai, Lin, Wang, Yuan, Ding, Zhang,
  Timofte, and Gool]{dauhst}
Yuanhao Cai, Jing Lin, Haoqian Wang, Xin Yuan, Henghui Ding, Yulun Zhang, Radu
  Timofte, and Luc~V Gool.
\newblock Degradation-aware unfolding half-shuffle transformer for spectral
  compressive imaging.
\newblock \emph{Advances in Neural Information Processing Systems},
  35:\penalty0 37749--37761, 2022{\natexlab{b}}.

\bibitem[Chen et~al.(2018{\natexlab{a}})Chen, Chen, Xu, and
  Koltun]{chen2018learning}
Chen Chen, Qifeng Chen, Jia Xu, and Vladlen Koltun.
\newblock Learning to see in the dark.
\newblock In \emph{CVPR}, 2018{\natexlab{a}}.

\bibitem[Chen et~al.(2018{\natexlab{b}})Chen, Chen, Xu, and
  Koltun]{chen2018sid}
Chen Chen, Qifeng Chen, Jia Xu, and Vladlen Koltun.
\newblock Learning to see in the dark.
\newblock In \emph{Proceedings of the IEEE/CVF Conference on Computer Vision
  and Pattern Recognition (CVPR)}, 2018{\natexlab{b}}.

\bibitem[Chu et~al.(2022)Chu, Chen, and Yu]{chu2022nafssr}
Xiaojie Chu, Liangyu Chen, and Wenqing Yu.
\newblock Nafssr: Stereo image super-resolution using nafnet.
\newblock In \emph{Proceedings of the IEEE/CVF Conference on Computer Vision
  and Pattern Recognition (CVPR) Workshops}, pages 1239--1248, 2022.

\bibitem[Dai et~al.(2023)Dai, Li, Zhou, Feng, Zhu, Sun, Sun, Loy, Gu, Liu,
  et~al.]{dai2023mipi}
Yuekun Dai, Chongyi Li, Shangchen Zhou, Ruicheng Feng, Qingpeng Zhu, Qianhui
  Sun, Wenxiu Sun, Chen~Change Loy, Jinwei Gu, Shuai Liu, et~al.
\newblock Mipi 2023 challenge on nighttime flare removal: Methods and results.
\newblock In \emph{Proceedings of the IEEE/CVF Conference on Computer Vision
  and Pattern Recognition}, pages 2852--2862, 2023.

\bibitem[Deng et~al.(2009)Deng, Dong, Socher, Li, Li, and Fei-Fei]{5206848}
Jia Deng, Wei Dong, Richard Socher, Li-Jia Li, Kai Li, and Li Fei-Fei.
\newblock Imagenet: A large-scale hierarchical image database.
\newblock In \emph{2009 IEEE Conference on Computer Vision and Pattern
  Recognition}, pages 248--255, 2009.

\bibitem[Dong et~al.(2023{\natexlab{a}})Dong, Gao, Li, Shi, and
  Liu]{dernn_lnlt}
Yubo Dong, Dahua Gao, Yuyan Li, Guangming Shi, and Danhua Liu.
\newblock Degradation estimation recurrent neural network with local and
  non-local priors for compressive spectral imaging.
\newblock \emph{arXiv preprint arXiv:2311.08808}, 2023{\natexlab{a}}.

\bibitem[Dong et~al.(2023{\natexlab{b}})Dong, Gao, Qiu, Li, Yang, and
  Shi]{rdluf_mixs2}
Yubo Dong, Dahua Gao, Tian Qiu, Yuyan Li, Minxi Yang, and Guangming Shi.
\newblock Residual degradation learning unfolding framework with mixing priors
  across spectral and spatial for compressive spectral imaging.
\newblock In \emph{Proceedings of the IEEE/CVF Conference on Computer Vision
  and Pattern Recognition}, pages 22262--22271, 2023{\natexlab{b}}.

\bibitem[Feng et~al.(2022{\natexlab{a}})Feng, Wang, Wang, and
  Huang]{Feng_Wang_Wang_Huang_2022}
Hansen Feng, Lizhi Wang, Yuzhi Wang, and Hua Huang.
\newblock Learnability enhancement for low-light raw denoising: Where paired
  real data meets noise modeling.
\newblock In \emph{Proceedings of the 30th ACM International Conference on
  Multimedia}, 2022{\natexlab{a}}.

\bibitem[Feng et~al.(2022{\natexlab{b}})Feng, Wang, Wang, and
  Huang]{feng2022learnability}
Hansen Feng, Lizhi Wang, Yuzhi Wang, and Hua Huang.
\newblock Learnability enhancement for low-light raw denoising: Where paired
  real data meets noise modeling.
\newblock In \emph{ACM MM}, 2022{\natexlab{b}}.

\bibitem[Jin et~al.(2023{\natexlab{a}})Jin, Han, Li, Guo, Chai, and
  Li]{jin2023dnf}
Xin Jin, Ling-Hao Han, Zhen Li, Chun-Le Guo, Zhi Chai, and Chongyi Li.
\newblock Dnf: Decouple and feedback network for seeing in the dark.
\newblock In \emph{CVPR}, 2023{\natexlab{a}}.

\bibitem[Jin et~al.(2023{\natexlab{b}})Jin, Xiao, Han, Guo, Zhang, Liu, and
  Li]{jiniccv23led}
Xin Jin, Jia-Wen Xiao, Ling-Hao Han, Chunle Guo, Ruixun Zhang, Xialei Liu, and
  Chongyi Li.
\newblock Lighting every darkness in two pairs: A calibration-free pipeline for
  raw denoising.
\newblock 2023{\natexlab{b}}.

\bibitem[Lehtinen et~al.(2018)Lehtinen, Munkberg, Hasselgren, Laine, Karras,
  Aittala, and Aila]{lehtinen2018noise2noise}
Jaakko Lehtinen, Jacob Munkberg, Jon Hasselgren, Samuli Laine, Tero Karras,
  Miika Aittala, and Timo Aila.
\newblock Noise2noise: Learning image restoration without clean data.
\newblock In \emph{CVPR}, 2018.

\bibitem[Liu et~al.(2022)Liu, Mao, Wu, Feichtenhofer, Darrell, and
  Xie]{liu2022convnet}
Zhuang Liu, Hanzi Mao, Chao-Yuan Wu, Christoph Feichtenhofer, Trevor Darrell,
  and Saining Xie.
\newblock A convnet for the 2020s, 2022.

\bibitem[Loshchilov and Hutter(2019)]{decoupled}
Ilya Loshchilov and Frank Hutter.
\newblock Decoupled weight decay regularization.
\newblock In \emph{International Conference on Learning Representations}, 2019.

\bibitem[Sun et~al.(2023{\natexlab{a}})Sun, Yang, Li, Zhou, Feng, Dai, Sun,
  Zhu, Loy, Gu, et~al.]{sun2023mipi}
Qianhui Sun, Qingyu Yang, Chongyi Li, Shangchen Zhou, Ruicheng Feng, Yuekun
  Dai, Wenxiu Sun, Qingpeng Zhu, Chen~Change Loy, Jinwei Gu, et~al.
\newblock Mipi 2023 challenge on rgbw remosaic: Methods and results.
\newblock In \emph{Proceedings of the IEEE/CVF Conference on Computer Vision
  and Pattern Recognition}, pages 2877--2884, 2023{\natexlab{a}}.

\bibitem[Sun et~al.(2023{\natexlab{b}})Sun, Yang, Li, Zhou, Feng, Dai, Sun,
  Zhu, Loy, Gu, et~al.]{sun2023mipi2}
Qianhui Sun, Qingyu Yang, Chongyi Li, Shangchen Zhou, Ruicheng Feng, Yuekun
  Dai, Wenxiu Sun, Qingpeng Zhu, Chen~Change Loy, Jinwei Gu, et~al.
\newblock Mipi 2023 challenge on rgbw fusion: Methods and results.
\newblock In \emph{Proceedings of the IEEE/CVF Conference on Computer Vision
  and Pattern Recognition}, pages 2870--2876, 2023{\natexlab{b}}.

\bibitem[Timofte et~al.(2016)Timofte, Rothe, and Van~Gool]{timofte2016seven}
Radu Timofte, Rasmus Rothe, and Luc Van~Gool.
\newblock Seven ways to improve example-based single image super resolution.
\newblock In \emph{Proceedings of the IEEE conference on computer vision and
  pattern recognition}, pages 1865--1873, 2016.

\bibitem[Ulyanov et~al.(2018)Ulyanov, Vedaldi, and Lempitsky]{ulyanov2018deep}
Dmitry Ulyanov, Andrea Vedaldi, and Victor Lempitsky.
\newblock Deep image prior.
\newblock In \emph{CVPR}, 2018.

\bibitem[Wang et~al.(2022)Wang, Cun, Bao, Zhou, Liu, and Li]{Uformer}
Zhendong Wang, Xiaodong Cun, Jianmin Bao, Wengang Zhou, Jianzhuang Liu, and
  Houqiang Li.
\newblock Uformer: A general u-shaped transformer for image restorationn.
\newblock In \emph{IEEE Conference on Computer Vision and Pattern Recognition},
  2022.

\bibitem[Wei et~al.(2021{\natexlab{a}})Wei, Fu, Zheng, and
  Yang]{Wei_Fu_Zheng_Yang_2021}
Kaixuan Wei, Ying Fu, Yinqiang Zheng, and Jiaolong Yang.
\newblock Physics-based noise modeling for extreme low-light photography.
\newblock page 1–1, 2021{\natexlab{a}}.

\bibitem[Wei et~al.(2021{\natexlab{b}})Wei, Fu, Zheng, and
  Yang]{wei2021physics}
Kaixuan Wei, Ying Fu, Yinqiang Zheng, and Jiaolong Yang.
\newblock Physics-based noise modeling for extreme low-light photography.
\newblock \emph{IEEE Transactions on Pattern Analysis and Machine
  Intelligence}, 2021{\natexlab{b}}.

\bibitem[Woo et~al.(2018)Woo, Park, Lee, and Kweon]{101007}
Sanghyun Woo, Jongchan Park, Joon-Young Lee, and In~So Kweon.
\newblock Cbam: Convolutional block attention module.
\newblock In \emph{Computer Vision -- ECCV 2018}, pages 3--19, Cham, 2018.
  Springer International Publishing.

\bibitem[Wu et~al.(2017)Wu, Duan, Liu, and Sun]{wu2017srpgan}
Bingzhe Wu, Haodong Duan, Zhichao Liu, and Guangyu Sun.
\newblock Srpgan: Perceptual generative adversarial network for single image
  super resolution, 2017.

\bibitem[Zamir et~al.(2022)Zamir, Arora, Khan, Hayat, Khan, and
  Yang]{zamir2022restormer}
Syed~Waqas Zamir, Aditya Arora, Salman Khan, Munawar Hayat, Fahad~Shahbaz Khan,
  and Ming-Hsuan Yang.
\newblock Restormer: Efficient transformer for high-resolution image
  restoration.
\newblock In \emph{Proceedings of the IEEE/CVF conference on computer vision
  and pattern recognition}, pages 5728--5739, 2022.

\bibitem[Zhang et~al.(2017)Zhang, Zuo, Chen, Meng, and Zhang]{zhang2017beyond}
Kai Zhang, Wangmeng Zuo, Yunjin Chen, Deyu Meng, and Lei Zhang.
\newblock Beyond a gaussian denoiser: Residual learning of deep cnn for image
  denoising.
\newblock \emph{IEEE TIP}, 2017.

\bibitem[Zhou et~al.(2019)Zhou, Siddiquee, Tajbakhsh, and
  Liang]{zhou2019unet++}
Zongwei Zhou, Md~Mahfuzur~Rahman Siddiquee, Nima Tajbakhsh, and Jianming Liang.
\newblock Unet++: Redesigning skip connections to exploit multiscale features
  in image segmentation.
\newblock pages 1856--1867. IEEE, 2019.

\bibitem[Zhu et~al.(2023)Zhu, Sun, Dai, Li, Zhou, Feng, Sun, Loy, Gu, Yu,
  et~al.]{zhu2023mipi}
Qingpeng Zhu, Wenxiu Sun, Yuekun Dai, Chongyi Li, Shangchen Zhou, Ruicheng
  Feng, Qianhui Sun, Chen~Change Loy, Jinwei Gu, Yi Yu, et~al.
\newblock Mipi 2023 challenge on rgb+ tof depth completion: Methods and
  results.
\newblock In \emph{Proceedings of the IEEE/CVF Conference on Computer Vision
  and Pattern Recognition}, pages 2863--2869, 2023.

\end{thebibliography}
}




\end{document}